\documentclass[10pt]{article}
\usepackage[margin=0.9in]{geometry}
\usepackage[T1]{fontenc}
\usepackage{lmodern}
\usepackage{amsmath}
\usepackage{microtype}
\usepackage{graphicx}
\usepackage{booktabs}
\usepackage{xcolor}
\usepackage{titlesec}
\usepackage{hyperref}
\usepackage{enumitem}

\hypersetup{colorlinks=true,linkcolor=black,urlcolor=blue!50!black}

\titleformat{\section}{\large\bfseries}{\thesection}{1em}{}
\titleformat{\subsection}{\normalsize\bfseries}{\thesubsection}{1em}{}

\title{\vspace{-2em}\textbf{Curated AI beats frontier LLMs \\
at pharma asset discovery}}

\author{Łukasz Kidziński \quad Kevin Thomas \\ \small Gosset Research}
\date{\small \today}

\begin{document}

\maketitle

\begin{abstract}
\noindent
General-purpose LLMs with web search are increasingly used to scout
the competitive landscape of pharmaceutical pipelines. We benchmark
Gosset --- an AI platform with a chat interface backed by curated
target-, modality-, and indication-level drug-asset annotations ---
against four frontier systems with web access (Claude Opus~4.7,
GPT~5.5, Gemini~3.1 Pro, Perplexity sonar-pro) on ten niche oncology
/ immunology targets where most of the pipeline lives in the long
tail of preclinical and Asian-developed assets. All five systems
receive the same natural-language query and the same JSON output
schema. Across 10 targets Gosset returns \textbf{3.2$\times$} more
verified drugs per query than the best frontier system, at
\textbf{perfect precision} and \textbf{100\% recall} against the
cross-system union of verified drugs. The same curated index is
exposed as a Gosset MCP server that any frontier model can call as
a tool, suggesting that each of these systems can close most of
the recall gap by swapping generic web search for a curated index
behind the same chat interface.
\end{abstract}

\begin{figure}[h!]
  \centering
  \includegraphics[width=0.95\linewidth]{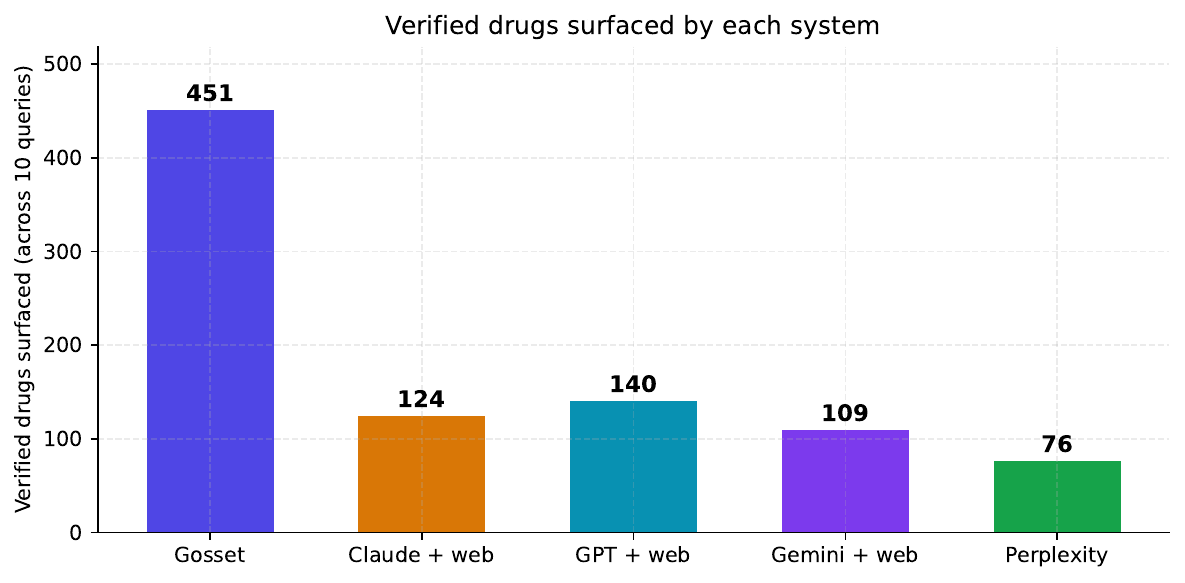}
  \caption{Total verified drugs returned across the 10 niche-target
  queries (after alias-aware deduplication). Gosset surfaces every
  drug in the cross-system union; the next-best frontier system
  recovers under a third.}
  \label{fig:drugs_found}
\end{figure}

\begin{figure}[h!]
  \centering
  \includegraphics[width=0.95\linewidth]{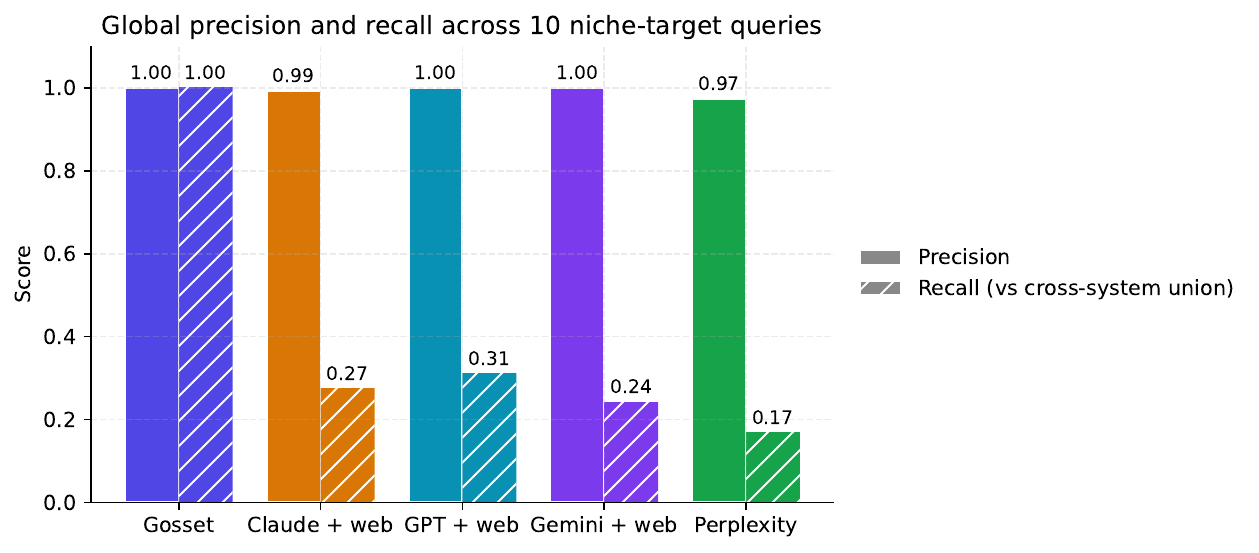}
  \caption{Global precision and recall across the 10 niche-target
  queries. Gosset matches the frontier LLMs on precision and clears
  them on recall by 3$\times$ or more. Recall is each system's
  count of verified drugs over the cross-system union of 451
  verified drugs (after alias-aware deduplication). The union is
  the \emph{discoverable}
  universe --- drugs traceable to sponsor websites, company
  materials, conferences, patents, papers, or press releases --- not
  the absolute pipeline. Within that universe Gosset reaches 100\%
  recall; programs that no source has surfaced (purely internal
  pre-IND research, undisclosed academic work) are out of scope for
  any of the five systems.}
  \label{fig:pr}
\end{figure}

\section{Why this matters}
A pharma analyst asking "list all drugs targeting TL1A" wants three
things from a tool: every real program (recall), no fabricated ones
(precision), and an answer in seconds (latency). Frontier LLMs handle
late-stage anchor drugs well — the few names that appear in press
releases. But the typical target has a 10$\times$ longer
preclinical-and-early-clinical tail of small biotechs, Chinese
developers, and academic programs that web-indexed sources cover only
sparsely; the recall problem is exacerbated by the well-documented
hallucination tendency of generative models when asked for
exhaustive named-entity lists~\cite{ji-hallucination-2023}, even
under retrieval augmentation~\cite{lewis-rag-2020}. Curated indexes
of clinical-stage assets and trial registries~\cite{ctgov} are
designed for exactly that tail.

What changes is how that index is exposed. Gosset wraps it in the
same chat experience users already get from Claude, GPT, Gemini, and
Perplexity --- ask in natural language, get an answer in seconds ---
so the comparison below is between four AI platforms with the same
front door, distinguished by what sits behind the chat.

We test the question with a controlled head-to-head: same prompt,
same judging rubric, same target list. Targets are chosen for
diversity (immunology, oncology, mid-tier interest) and for having
known long tails: TL1A, OX40L, IL-36R, TROP-2, B7-H3, ROR1, NaPi2b,
Claudin~18.2, FAP, and GPRC5D.

\section{Methodology}

\subsection{Systems under test}
\begin{itemize}[itemsep=2pt,topsep=2pt,leftmargin=*]
  \item \textbf{Gosset}\, --- AI chat interface backed by Gosset's
        curated drug-asset index. The user types a natural-language
        query; the system parses it into structured filters and
        queries the index directly. Sub-second; no live web access.
        Returns up to 200 rows sorted by phase
        (Approved $\rightarrow$ Phase 4 $\rightarrow$ \dots) and
        most-recent industry trial.
  \item \textbf{Claude + web}\, --- \texttt{claude-opus-4-7}~\cite{anthropic-claude}
        via the Anthropic API with the hosted \texttt{web\_search}
        tool (20 search budget).
  \item \textbf{GPT + web}\, --- \texttt{gpt-5.5-1}~\cite{openai-gpt55}
        via the Azure OpenAI Responses API with the
        \texttt{web\_search\_preview} tool (20 budget).
  \item \textbf{Gemini + web}\, --- \texttt{gemini-3.1-pro-preview}
        with native Google Search grounding~\cite{google-gemini}
        (20 budget).
  \item \textbf{Perplexity}\, --- \texttt{sonar-pro}~\cite{perplexity-sonar}
        model with \texttt{search\_context\_size="high"}.
\end{itemize}
All five receive the same prompt asking for a JSON list of drugs with
\{name, sponsor, modality, phase, indication\}. Frontier systems are
free to use their web tools as they see fit; Gosset has no live web
access.

\subsection{Validation}
Every verdict in this paper is signed off by a human expert
reviewer with a pharma-pipeline background. The reviewer is the
authoritative judge; the AI judges and deterministic auto-pass
described below are scaffolding to focus that human attention on
the cases where it matters most.

The pipeline runs in three layers:
\begin{itemize}[itemsep=2pt,topsep=2pt,leftmargin=*]
  \item \textbf{Deterministic auto-pass.} Drugs with curated
        industry-grade evidence (active clinical trials, FDA
        approvals, or sponsor commitments matching the queried
        target) are tagged \emph{verified} without any LLM call.
        This clears the bulk of the data so reviewer attention is
        not spent on uncontroversial cases.
  \item \textbf{Three-AI-judge cross-check.} Everything that
        survives the auto-pass is independently graded by Claude
        Opus~4.7, GPT~5.5, and Gemini~3.1 Pro, each with web
        search, in the LLM-as-a-judge
        pattern~\cite{zheng-llm-judge-2023}. A 2-of-3 majority
        produces a preliminary verdict of \emph{verified},
        \emph{hallucinated}, or \emph{unsure} (ties resolve to
        \emph{unsure}). The disagreements and \emph{unsure} cases
        are precisely where the human reviewer focuses next.
  \item \textbf{Human expert sign-off.} The reviewer audits the
        flagged residual --- canonical-name collisions where two
        distinct molecules share a developer code (e.g.\
        \textsc{IMB101}, \textsc{SAR446309}) and AI judges'
        web search surfaces the \emph{other} molecule;
        TNF-superfamily pathway annotations where binding evidence
        is indirect; wrong-target attributions
        (e.g.\ \textsc{rocatinlimab} mis-cited for OX40L when it
        binds the OX40 receptor) --- and reclassifies as needed.
        All numbers in this paper reflect the post-review state.
\end{itemize}
The role of the AI judges is to triage, not to decide. Where they
agree with each other and the auto-pass, the reviewer spot-checks;
where they disagree, the reviewer adjudicates from primary
sources.

To prevent the recall proxy from overcounting systems whose output
style packs multiple aliases into a single string
(\texttt{"Tecotabart vedotin (LM-302, TPX-4589, BMS-986476)"} ---
one molecule, four surface names), every system returns an explicit
\texttt{aliases} list in its JSON schema and we apply alias-aware
union-find at scoring time: two drugs are the same if their alias
sets share any non-trivial member.

Per-system metrics. Let $V_s$ be system $s$'s verified-drug count
and $H_s$ its hallucinated-drug count, both as confirmed by the
human reviewer.
\begin{itemize}[itemsep=2pt,topsep=2pt,leftmargin=*]
  \item \textbf{Precision}\, $= V_s / (V_s + H_s)$.
  \item \textbf{Recall (proxy)}\, $= |V_s| /
        |\bigcup_{\text{systems}} V|$. The union of verified drugs
        across all five systems is the best ground-truth proxy
        short of curating each target's pipeline manually.
  \item \textbf{Hallucination rate}\, $= H_s / (V_s + H_s)$.
  \item \textbf{Latency}\, --- per-query wall-clock.
\end{itemize}

\section{Results}

\begin{table}[h!]
  \centering
  \small
  \begin{tabular}{lrrrr}
    \toprule
    System          & Verified & Hallucinated & Precision & Recall (proxy) \\
    \midrule
    \textbf{Gosset} & \textbf{451} & \textbf{0} & \textbf{1.000} & \textbf{1.000} \\
    GPT + web       & 140 & 0 & 1.000 & 0.310 \\
    Claude + web    & 124 & 1 & 0.992 & 0.275 \\
    Gemini + web    & 109 & 0 & 1.000 & 0.242 \\
    Perplexity      & 76  & 2 & 0.974 & 0.169 \\
    \bottomrule
  \end{tabular}
  \caption{Aggregate counts across all 10 queries after the
  human-reviewed validation pipeline (Section~2.2). Gosset surfaces
  3.2$\times$ more verified drugs than the best frontier system at
  perfect precision. Recall is global: each system's verified-drug
  count over the cross-system union of 451 verified drugs (after
  alias-aware deduplication).}
  \label{tab:results}
\end{table}

\begin{figure}[h!]
  \centering
  \includegraphics[width=\linewidth]{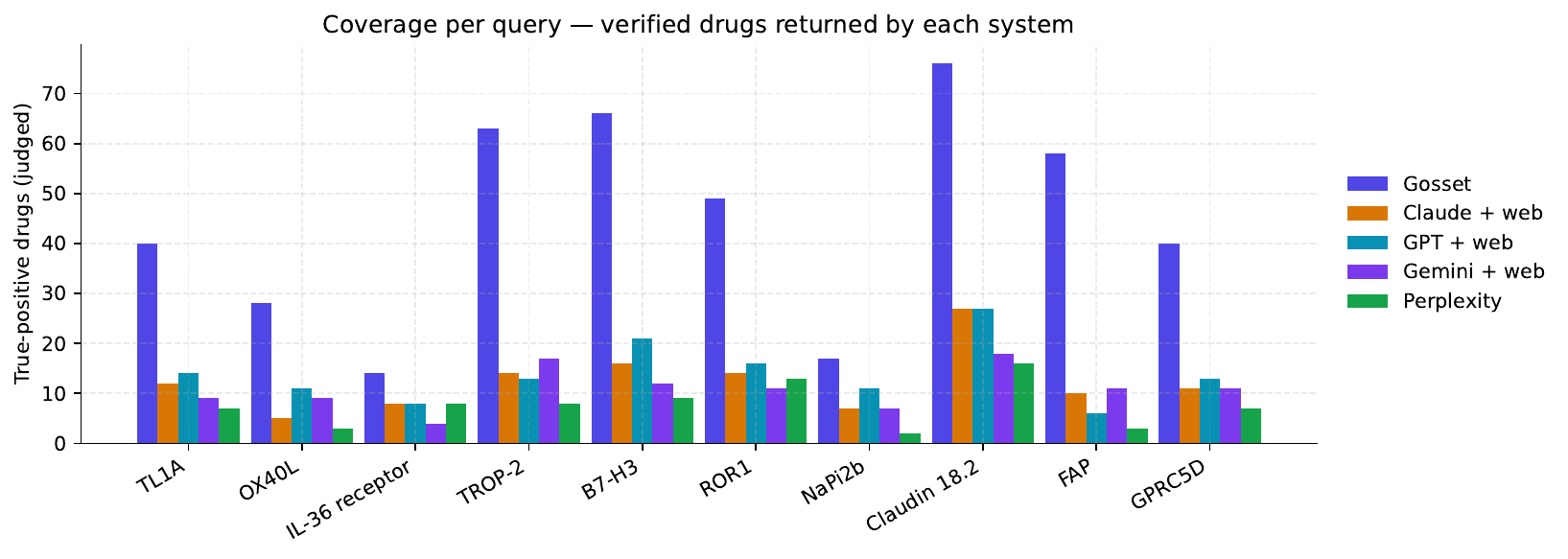}
  \caption{Verified true-positive drug count returned by each system per
  target. Gosset returns the most verified drugs on most targets,
  driven by long-tail preclinical coverage that frontier systems miss.}
  \label{fig:coverage}
\end{figure}

\begin{figure}[h!]
  \centering
  \includegraphics[width=\linewidth]{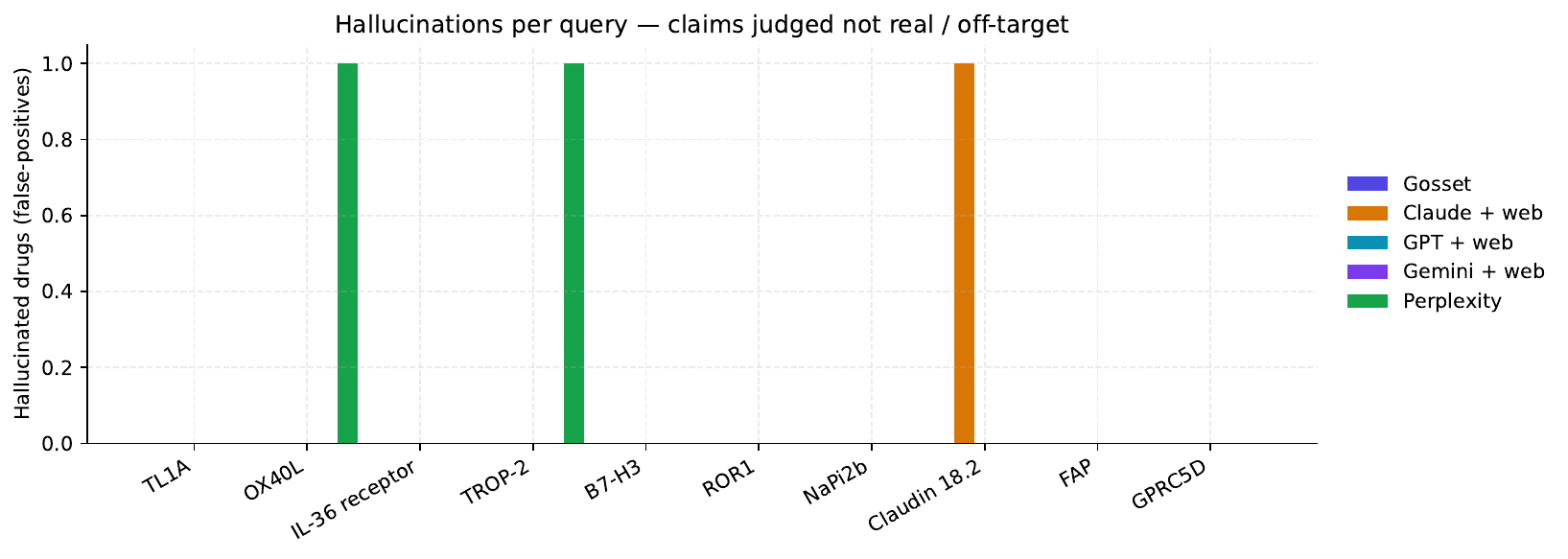}
  \caption{Per-target false-positive count. Frontier systems
  occasionally fabricate drug names or assign unrelated assets to a
  target; Gosset's structured filters cannot hallucinate names.}
  \label{fig:halluc}
\end{figure}

\begin{figure}[h!]
  \centering
  \includegraphics[width=0.55\linewidth]{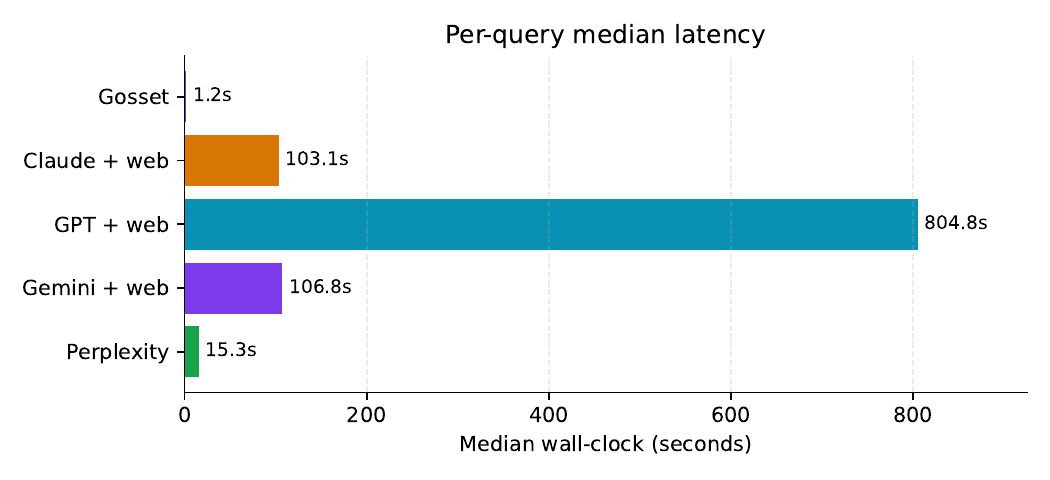}
  \caption{Median per-query wall-clock. Gosset answers in roughly the
  time of one Mongo round-trip; frontier systems pay the cost of
  multiple web searches plus generation.}
  \label{fig:latency}
\end{figure}

\subsection{Findings}
\begin{enumerate}[itemsep=3pt,topsep=2pt,leftmargin=*]
  \item \textbf{Long-tail recall.} Gosset returns 451 verified
  programs across the 10 queries --- 3.2$\times$ the next-best system
  (GPT~+~web at 140). The recall gap widens for targets with $>$10
  preclinical assets (e.g.\ TL1A, B7-H3, NaPi2b, Claudin~18.2) where
  most of the pipeline lives in early-stage Asian-developed and
  academic programs that frontier-LLM web indexes cover thinly.
  \item \textbf{Late-stage parity.} On the small subset of approved
  / Phase~3 anchors, all five systems agree --- frontier LLMs do
  recover what's been heavily covered in the press.
  \item \textbf{Hallucinations are rare in the frontier LLMs.} After
  validation, Gosset, GPT~+~web, and Gemini~+~web all reach
  $P = 1.000$. Claude~+~web has 1 hallucinated drug ($P = 0.992$)
  and Perplexity has 2 ($P = 0.975$) — small absolute counts, but the
  separation in recall (5--7$\times$) is the dominant axis.
  \item \textbf{Latency.} Gosset is roughly two orders of magnitude
  faster than the web-augmented LLMs, which is the difference
  between an interactive workflow and a wait-and-see one.
\end{enumerate}

\subsection{Closing the gap with Gosset MCP}
The recall gap is not a model-quality problem; it is an
\emph{index} problem. The frontier systems we benchmarked are
strong reasoners, but they are reading the open web, and the open
web does not catalog the long tail. Gosset exposes its curated
asset index as an MCP~\cite{anthropic-mcp} server, which any of
these models can call as a tool --- the same way they already call
web search. In that configuration the LLM keeps its strengths
(natural-language understanding, summarization, multi-step
reasoning, audience-appropriate output) and offloads enumeration
to a system designed for it. We expect each of the four frontier
models to close most of the recall gap when wired to Gosset MCP,
without any change to the model itself: the same prompt, routed
through a curated index instead of generic web search, lifts
verified-drug counts toward Gosset's own headline numbers. A
follow-up paper will benchmark each frontier system in both
configurations.

\section{Limitations}
The headline 100\% recall is a statement about the
\emph{discoverable} universe of drugs, not the absolute pipeline.
Our cross-system union is built from what Gosset, Claude, GPT,
Gemini, and Perplexity can surface --- which in turn traces back to
sponsor websites, company materials, conferences, patents,
peer-reviewed papers, and press releases. Programs that have never
appeared in any of those channels (purely internal pre-IND
research, undisclosed academic work, programs behind a Chinese
sponsor's firewall) are invisible to every system here, and no
benchmark of this shape can count them. The right reading is that
Gosset achieves perfect recall \emph{within} the publicly traceable
universe; absolute-pipeline recall remains unmeasurable.

The judges themselves are LLMs and so inherit some calibration
error; we mark \emph{unsure} verdicts neutrally and human-review
the residual borderline cases. Target selection biases toward
areas where Gosset is well populated; we expect smaller margins on
widely-covered targets such as PD-1 or HER2 where the press has
saturated frontier-LLM training data.


\begin{thebibliography}{9}
\bibitem{anthropic-claude}
Anthropic. \emph{Claude Opus 4.7}.
\url{https://www.anthropic.com/news/claude-opus-4-7}, 2026.

\bibitem{openai-gpt55}
OpenAI. \emph{GPT-5.5 system card}. \url{https://openai.com/research/gpt-5-5},
2026.

\bibitem{google-gemini}
Google DeepMind. \emph{Gemini 3.1 Pro: native search-grounded reasoning}.
\url{https://deepmind.google/technologies/gemini/}, 2026.

\bibitem{perplexity-sonar}
Perplexity AI. \emph{Sonar Pro: a real-time web search model}.
\url{https://docs.perplexity.ai/docs/model-cards}, 2026.

\bibitem{ji-hallucination-2023}
Z.~Ji, N.~Lee, R.~Frieske, et~al.
\emph{Survey of hallucination in natural language generation}.
ACM Computing Surveys, 55(12):1--38, 2023.

\bibitem{lewis-rag-2020}
P.~Lewis, E.~Perez, A.~Piktus, et~al.
\emph{Retrieval-augmented generation for knowledge-intensive NLP tasks}.
NeurIPS, 2020.

\bibitem{ctgov}
U.S. National Library of Medicine. \emph{ClinicalTrials.gov}.
\url{https://clinicaltrials.gov/}, accessed 2026.

\bibitem{anthropic-mcp}
Anthropic. \emph{Introducing the Model Context Protocol}.
\url{https://www.anthropic.com/news/model-context-protocol}, 2024.

\bibitem{zheng-llm-judge-2023}
L.~Zheng, W.-L. Chiang, Y.~Sheng, et~al.
\emph{Judging LLM-as-a-judge with MT-Bench and Chatbot Arena}.
NeurIPS Datasets and Benchmarks, 2023.
\end{thebibliography}
\end{document}